\documentclass{article} 
\usepackage{iclr2021_conference,times}
\usepackage{float}

\usepackage{amsmath,amsfonts,bm}

\def\eqref#1{equation~\ref{#1}}

\def\1{\bm{1}}

\DeclareMathAlphabet{\mathsfit}{\encodingdefault}{\sfdefault}{m}{sl}
\SetMathAlphabet{\mathsfit}{bold}{\encodingdefault}{\sfdefault}{bx}{n}

\usepackage{graphicx}

\usepackage{hyperref}
\usepackage{url}
\usepackage[utf8]{inputenc} 
\usepackage[T1]{fontenc}    
\usepackage{booktabs}       
\usepackage{amsfonts}       
\usepackage{nicefrac}       
\usepackage{microtype}      
\usepackage{xcolor}         

\usepackage[none]{hyphenat}
\usepackage{graphicx}
\usepackage{multirow}
\usepackage{csquotes}
\usepackage{subcaption}
\usepackage{lineno}
\usepackage{hyperref}
\usepackage{url}
\usepackage{multirow}
\usepackage{graphicx}
\makeatletter
\def\namedlabel#1#2{\begingroup
    #2%
    \def\@currentlabel{#2}%
    \phantomsection\label{#1}\endgroup
}
\makeatother

\title{Empirical evaluation of \mbox{Uncertainty Quantification} in Retrieval-Augmented Language Models \\ for Science}

\author{Sridevi Wagle, Sai Munikoti, Anurag Acharya, Sara Smith \& Sameera Horawalavithana \\
Pacific Northwest National Laboratory\\
Richland, WA, USA \\
\texttt{\{sridevi.wagle,sai.munikoti,anurag.acharya,sara.smith,}\\
\texttt{yasanka.horawalavithana\}@pnnl.gov} \\
}

\iclrfinalcopy 
\begin{document}

\maketitle

\begin{abstract}

Large language models (LLMs) have shown remarkable achievements in natural language processing tasks, producing high-quality outputs. However, LLMs still exhibit limitations, including the generation of factually incorrect information. In safety-critical applications, it is important to assess the confidence of LLM-generated content to make informed decisions. Retrieval Augmented Language Models (RALMs) is relatively a new area of research in Natural Language Processing (NLP). RALMs offer potential benefits for scientific NLP tasks, as retrieved documents, can serve as evidence to support model-generated content. This inclusion of evidence enhances trustworthiness, as users can verify and explore the retrieved documents to validate model outputs. Quantifying uncertainty in RALM generations further improves trustworthiness, with retrieved text and confidence scores contributing to a comprehensive and reliable model for scientific applications. 
However, there is limited to no research on UQ for RALMs, particularly in scientific contexts. This study aims to address this gap by conducting a comprehensive evaluation of UQ in RALMs, focusing on scientific tasks. 
This research investigates how uncertainty scores vary when scientific knowledge is incorporated as pretraining and retrieval data and explores the relationship between uncertainty scores and the accuracy of model-generated outputs.
We observe that an existing RALM finetuned with scientific knowledge as the retrieval data tends to be more confident in generating predictions compared to the model pretrained only with scientific knowledge.
We also found that RALMs are overconfident in their predictions, making inaccurate predictions more confidently than accurate ones. 
Scientific knowledge provided either as pretraining or retrieval corpus does not help alleviate this issue.
We released our code, data and dashboards at \url{https://github.com/pnnl/EXPERT2}.
\end{abstract}

\section{Introduction}


The continuous development of various Large Language Models (LLMs) has achieved a human level accuracy on various natural language processing tasks including machine translation, question answering and code generation. The rapid progress in this area has been driven by the availability of large datasets of text and code, as well as advances in computing power and machine learning algorithms. Some of the most notable LLMs include GPT-3 \citep{brown2020language}, LaMDA \citep{thoppilan2022lamda}, Codex \citep{finnie2022robots} and Megatron-Turing NLG \citep{smith2022using}.
These have been trained on datasets of hundreds of billions or even trillions of words, and can generate text that is often indistinguishable from human-written text. However, the models can still provide factually incorrect answers or what is commonly known as ``hallucinations''. Therefore, it is crucial to access the model confidence on its generation for informed decisions in safety critical applications.

There are standard UQ approaches in machine learning which are extended to LLMs. Notable among them are Temperature scaling/calibration \citep{xiao2022uncertainty}, Ensembles via Monte-carlo dropout \citep{gal2016dropout}, Last-Layer Stochastic Variational Inference (LL SVI )~\citep{lakshminarayanan2017simple}. However, most of the existing approaches are not valid for generative LLMs. 
Therefore,~\citet{kuhn2023semantic} proposed semantic entropy to accurately assess the uncertainty in generative models in a computational efficient manner. 



Furthermore, retrieval augmented language models (RALMs) are an advancement on top of regular LLMs that have become increasingly popular~\citep{izacard2022few, munikoti2023evaluating}. 
This is mainly due to their grounding capability and adaptability to work with new data sets and being small in size by decentralizing learning from parameters to external knowledge store.  RALMs are being used in a variety of applications, including Google Search, Bard \citep{thoppilan2022lamda}, and GPT-3 Retriever \citep{brown2020language, karpukhin2020dense}. These applications use RALMs to generate snippets of text, answer questions, and generate creative content. RALM consists of two major components, namely Retriever and Reader. Retriever retrieves documents (text chunks) from the corpus and Reader (Language model) uses the retrieved documents as extra context to support its generation.
RALMs appear promising for scientific NLP applications because the retrieved scientific texts during generation can serve as supporting evidence for the model's outputs.
Evidence instills trustworthiness in the model since
experts can further examine the retrieved documents to validate the model's reasoning and conclusions.




The trustworthiness of RALM can further be improved by quantifying uncertainty in RALM generations.
The combination of retrieved scientific text along with uncertainty estimates provides a comprehensive and reliable model suited for practical usage in the scientific domain. 
In this study, we experiment with different uncertainty measures to assess the reliability of RALM using three scientific benchmark datasets. In this regard, we choose ATLAS \citep{izacard2022few} model which is a state-of-the-art RALM. 
Our objective is to empirically analyze how uncertainty scores vary when scientific knowledge is used either as the pretraining or retrieval data. 
We conduct a series of experiments with a scientific document corpus to train two RALMs, where one model uses the scientific document corpus in both Retriever and Reader while another model is only finetuned with the scientific document corpus provided as a retrieval corpus in the Retriever.

\section{Related work}

There has been significant amount of work in the field of uncertainty quantification for deep neural networks. \citet{xiao2022uncertainty} mentions different quantifiers for uncertainty. Temperature Scaling is one such technique that is used to quantify uncertainty in machine learning model predictions by providing a measure of how confident the model is in its predictions. 
\citet{gal2016dropout} propose the Monte Carlo Dropout technique to estimate the uncertainty of machine learning model predictions. They do so by normalizing the results of multiple forward passes through the model with dropout turned on. 
It is simple to implement and can be used with a variety of different model architectures. However, it is expensive for very large scale models such as LLMs.
Other uncertainty quantifiers include Ensemble and LL SVI (Last-Layer Stochastic Variational Inference). Ensemble learning works by combining the predictions of multiple independently trained models \citep{lakshminarayanan2017simple}, while LL SVI uses variational inference to approximate the posterior predictive distribution of the model \citep{blundell2015weight}. Both techniques can be effective for improving the reliability and robustness of machine learning models.

However, despite this host of works in the field of UQ, there has been limited number of works when it comes to quantifying uncertainty of generative large language models (LLMs). 
More recently, however, uncertainty quantification for large language models(LLMs) has become a key area of research, as there has been increased interest in employing LLMs in real-world critical applications. 
\citet{lin2023generating} propose a number of methods for quantifying the uncertainty of LLMs, using the entropy of the distribution of generated tokens, the diversity of the generated responses, and the average semantic dispersion of the generated responses. Their methods are simple to implement and can be used to improve the reliability of LLMs. 
\citet{jiang2021can} experiment with different techniques to improve the reliability of language models by making them aware of their own limitations by developing methods to calibrate the confidence scores of LMs so that they better reflect the actual probability of correctness. They develop several methods for calibrating LMs, which have shown to be effective on different datasets, but also have limitations since they may not work well for different types of LMs or tasks.
\citet{kadavath2022language} show that 
self-evaluation can be approached on open-ended tasks by first asking models to propose answers, and then asking them to evaluate the probability that their answers are correct. 
They also investigate whether models can be trained to predict the probability that they know the answer to a question, without reference to any particular proposed answer.
Models perform well at predicting this probability, and partially generalize across tasks. However, they struggle to calibrate their predictions on new tasks.
\citet{kuhn2023semantic} use a relatively new method of Semantic Entropy for estimating uncertainty in natural language generation (NLG), which is a measure of the uncertainty of a generated text sequence that uses linguistic invariances. They evaluate this measure on two NLG datasets, TriviaQA \citep{joshi-etal-2017-triviaqa}  and CoQA \citep{reddy2019coqa}, and find that it outperforms previous methods on both datasets. The authors' experiments show that semantic entropy is more accurate and informative than previous methods, and that it can be used to improve the performance of NLG models on downstream tasks.

Despite these works, there is a lack of research in the area of uncertainty quantification for RALM both in general as well as science focused tasks. Therefore, in this work, we conduct comprehensive evaluation of UQ in RALMs for science tasks.

\section{Methodology} \label{methodology}
In this section, we discuss RALMs and uncertainty measures used in our empirical study.

\subsection{Retrieval augmented language models}
\begin{figure}[htbp]
    \centering
    \includegraphics[scale=0.4]{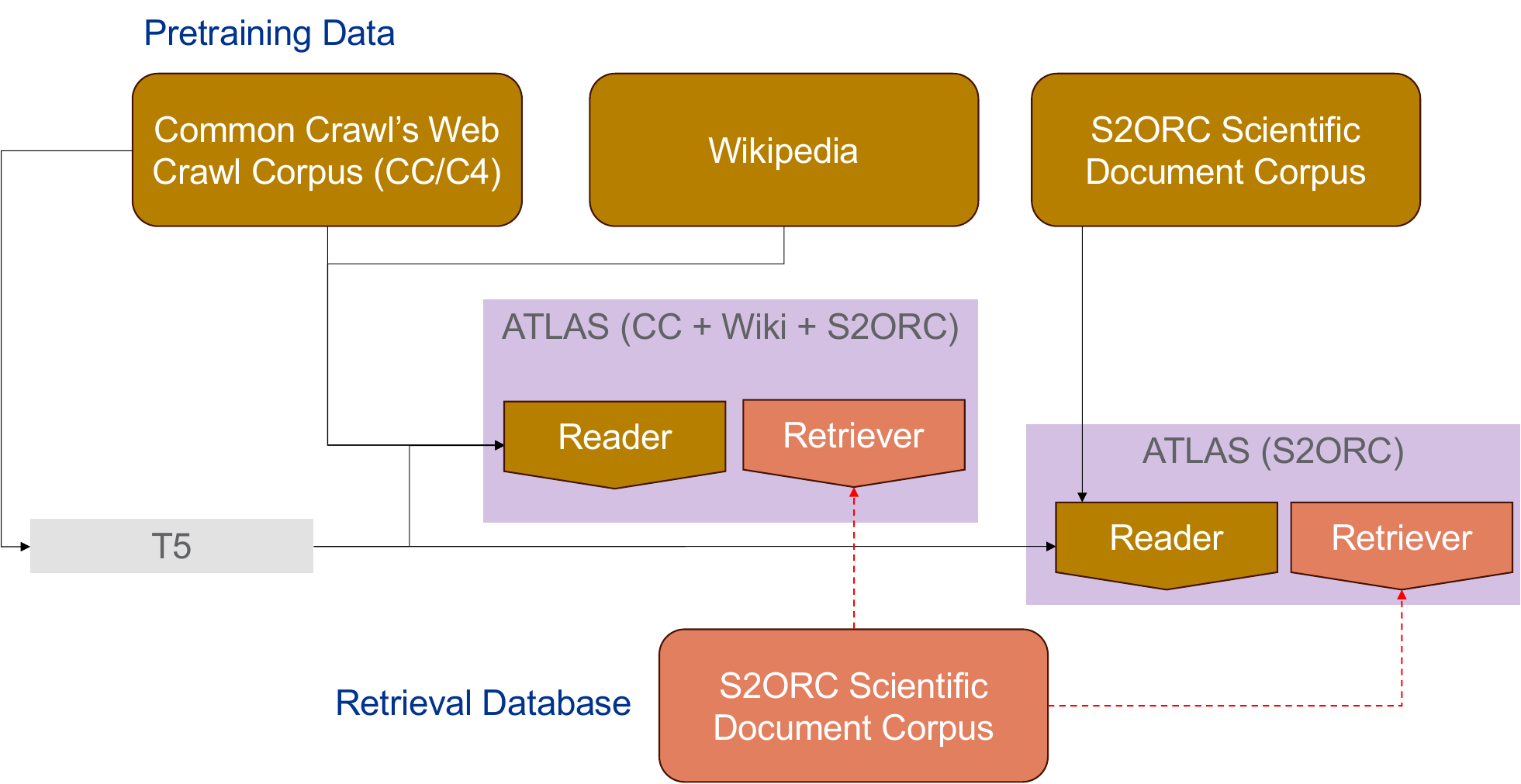}
    \caption{ATLAS Model Variants}
    \label{fig:atlas-Model_variants}
\end{figure}

Retrieval augmented language model 
(RALM) consists of two major components, Retriever and Reader. Retriever retrieves relevant documents from the text corpus and use them as context to support Reader (i.e., language model) to generate output (Figure~\ref{fig:atlas-Model_variants}). 
We choose RALM over other types of LLM for science tasks because the retrieved scientific texts can be used as evidences to instill trust in model predictions and enable grounding which is critical in scientific applications. There are various works in this space, such as REALM \citep{guu2020retrieval}, DSP \citep{khattab2022demonstrate}, ATLAS \citep{izacard2022few}, and so on. We choose the ATLAS model to conduct all our experiments described in this paper.

ATLAS is specifically designed to work for knowledge intensive tasks (e.g., fact checking question answering, etc.) in few-shot settings. ATLAS retrieves relevant documents based on the input query by using a dense retriever based on the Contriever \citep{izacard2021unsupervised}. The retrieved documents are processed along with the input query to a T5 model using the Fusion-in-Decoder architecture that generates the corresponding output \citep{izacard2020leveraging}. We prefer ATLAS for science tasks because it (i) allows end to end training of retriever and language model (enables us to pretrain model with scientific data), (ii) offers various techniques/configurations (e.g., query side finetuning) for efficient training and evaluation, and (iii) has vector database and few-shot ability enables seamless adaptation to several domains (Humanities, Social Sciences, STEM, etc) in science tasks.

\subsection{UQ measures for Retrieval augmented Language models}
Calibration is a widely used UQ approach for large scale deep neural network (DNN). This is due to its low computational complexity compared to other UQ approaches such as Ensemble and MC Dropout. In calibration, logits from the outermost layer of a DNN are normalized to provide confidence scores such that expected accuracy matches confidence scores \citep{guo2017calibration}. Calibration has been used for various discriminative tasks in LLM \citep{wang2022uncertainty,si2022revisiting}, including text classification, entity recognition. However, it is not suitable for generative tasks in LLM due to variable output length. Therefore, several approximation UQ measures have been proposed recently in the literature.

For this work, we use two of those measures to perform the uncertainty quantification,~\textit{Normalized Predictive Entropy} and~\textit{Semantic Entropy}, which are described in detail in the following subsections.

\subsubsection{Normalized Predictive Entropy} 
The entropy is a statistical parameter which measures, in a certain sense, how much information is produced on the average for each letter of a text in the language. Let us assume the RALM generates M output sequence and sequence $s$ consists of $N$ tokens. The entropy of the sequence $s$ is the sum of product of conditional probabilities of all tokens in $s$ and their corresponding log values. 
\begin{equation}
  E(s) = \sum_{i=0} ^{N} P(s_i|s_{<i}, x)logP(s_i|s_{<i}, x)   
\end{equation}
where $s_{i}$ denotes $i^{th}$ token of sequence $s$ and $s_{<i}$ represents all tokens upto $i^{th}$ position.

To calculate the net predictive entropy of the model for a given query (prompt) $x$, we average the predictive entropy across all $M$ generated sequences in the set $S$ as shown below:
\begin{equation}
    \begin{split}
     PE(x) & = \frac{1}{M}\sum_{s \in S} E(s) \\
     & =  \frac{1}{M}\sum_{s \in S} \sum_{i=0} ^{N} P(s_i|s_{<i}, x)logP(s_i|s_{<i}, x)  
     \end{split}
     \label{eqn:2}
\end{equation}

Generated sequences can be of different length. Therefore, as for eqn. \ref{eqn:2}, the longer
sequences have lower joint likelihoods because of the conditional independence of the token probabilities. Hence, negative log-probability of a sequence grows linearly with the length of the sequence, so longer sequences tend to contribute more to predictive entropy \citep{kuhn2023semantic}. To counter this effect, the log-probabilities are normalized by the length of the sequence while calculating sequence entropy. The normalized form of predictive entropy can be expressed as: 
\begin{equation} \label{norm_en}
NPE(x) =  \frac{1}{M}\sum_{s \in S} \frac{1}{N_{s}} \sum_{i=0} ^{N_{s}} P(s_i|s_{<i}, x)logP(s_i|s_{<i}, x) ,
\end{equation}
where $N_{s}$ denotes the length of the sequence $s$.

\subsubsection{Semantic Entropy}
Predictive entropy as we discussed in the above subsection, approximates model uncertainty via token wise likelihoods (i.e. lexical confidence). However, in free form text generation, one always cares about the semantic meaning of the entire generated sequence. For instance, predictive entropy could be high if a model is uncertain about whether to
generate “Japan’s capital is Tokyo” or “Tokyo is Japan’s capital”. However model's uncertainty is actually low in this example since both sequence are semantically equiavalent \citep{kuhn2023semantic}. To address this shortcoming of predictive entropy, \citet{kuhn2023semantic} introduces new UQ metric, \textit{semantic entropy}. To this end, the authors 
compute semantic likelihoods — probabilities attached to meanings
of generated sequence rather than standard sequence-likelihoods. A clustering algorithm is implemented so that sequences with similar meaning group together. Then, semantic-likelihoods are computed for each meanings set (cluster) rather than each sequence. 


Lets assume we obtain a finite number of meaning sets, $C$. We can use the sum of these sets to calculate the Semantic Entropy similar to the way we calculate predictive entropy for a sequence.

\begin{equation} \label{sem_en}
    SE(x) = -\sum_{c}\left(\left(\sum_{s \in c}P(s|x)\right)\log\left(\sum_{s \in c}P(s|x)\right)\right)
\end{equation}
Since one cannot have all possible meaning class-$c$ in the limited number of generations, we need to take the estimation in eqn. \ref{sem_en} to get the semantic entropy as shown below:
\begin{equation} \label{sem_en2}
    SE(x) \approx - \frac{1}{|C|}
    \sum_{i=1}^{|C|} \log p (C_{i}|x)= - \frac{1}{|C|}
    \sum_{i=1}^{|C|} \log \left(\sum_{s \in c_{i}}P(s|x)\right),
\end{equation}
where $C_{i}$ belongs to cluster (meaning set) $i$.

We prefer Normalized entropy and Semantic entropy over other UQ measures because normalized entropy performs token wise quantification and semantic entropy captures the semantic meaning, thus covering multiple UQ aspects for RALM.





\begin{table*}[H]
\centering
\caption{Summary of different pretraining, instruction tuning and benchmark datasets}

\label{tab:model_setup}
\scalebox{0.8}{
\begin{tabular}{|c|c|c|c|c|c|c|}
\hline
\multirow{2}{*}{Model}                      & \multicolumn{2}{c|}{Pretraining}                                         \\ \cline{2-3} 
                                            & \multicolumn{1}{c|}{Data}                       & Retrieval Corpus       \\ \hline
\multirow{2}{*}{ATLAS   (CC+ Wiki + S2ORC)} & \multicolumn{1}{c|}{\multirow{2}{*}{CC + Wiki}} & \multirow{2}{*}{Wiki}  \\
                                            & \multicolumn{1}{c|}{}                           &                        \\ \hline
\multirow{2}{*}{ATLAS (S2ORC)}              & \multicolumn{1}{c|}{\multirow{2}{*}{S2ORC}}     & \multirow{2}{*}{S2ORC} \\
                                            & \multicolumn{1}{c|}{}                           &                        \\ \hline
\end{tabular}
}
\end{table*}


\section{Experimental Setup}

In this section, we describe the datasets (Section~\ref{datasets}), and model variants (Section~\ref{sec:models}) used in our experiments.
We also release our implementation and Human-AI reasoning dashboards (see Figure~\ref{fig:uq_widget} in Appendix~\ref{appendix_widgets}) used to conduct experiments publicly\footnote{\url{https://github.com/pnnl/EXPERT2}}.

\subsection{Benchmarks and Datasets} 
\label{datasets}
We leverage two types of scientific benchmarks for the uncertainty evaluation of RALMs. The first is SciRepEval, which offers $25$ tasks spanning classification, regression, ranking, and search formats. Specifically, we focus on the classification tasks of Fields of Study (FoS) and MAG because they will test the ability of the models to recognize diverse scientific domains. The second benchmark we use is MMLU, which provides 57 multi-choice question-answering datasets retrieved from real-world examinations across diverse scientific fields. It is categorized into four major subdomains - humanities, social sciences, STEM, and other. This benchmark tests the ability of the models to understand the diverse science context for accurate generations.

We use the S2ORC~\citep{lo2019s2orc} data which is a large corpus of curated $31.1M$ English-language academic papers spanning many academic disciplines. 
We preprocess the S2ORC~\citep{lo2019s2orc} dataset to create a corpus of $354M$ text passages.
Each passage has a maximum of $512$ tokens, or $100$ words, that are concatenated with the corresponding title of the document the passage belongs to. Our text corpus spans across $19$ different scientific domains.

\subsection{Atlas Model Variants}
\label{sec:models}

We analyze uncertainty results across two ATLAS model variants as shown in Figure~\ref{fig:atlas-Model_variants}. 

\paragraph{ATLAS (CC + Wiki + S2ORC):} ATLAS model combines autoregressive text generation with retrieval-based language model pretraining based on the encoder-decoder architecture and fine-tuned on open-domainQA \citep{izacard2022few}. ATLAS (CC + Wiki + S2ORC) uses the Fusion-in-decoder architecture to fuse the retrieved text chunks with the input queries during the pretraining. In this configuration, ATLAS model is pretrained with a Common Crawl  (CC) and Wikipedia. The S2ORC dataset is a retrieval corpus in instruction fine-tuning and evaluation.
All our experiments are based on ATLAS base model with $220M$ parameters.

\paragraph{ATLAS (S2ORC):} 
We train the ATLAS (S2ORC) model from scratch with the S2ORC  scientific text datasets. For a fair comparison with ATLAS (CC + Wiki + S2ORC), we initialize the ATLAS (S2ORC) model with the T5-lm-adapt \citep{raffel2020exploring} model and trained jointly with the retrieval model, Contriever \citep{izacard2021unsupervised}. We encode the scientific text passages (354M) with the Contriever model and construct a document index in the FLAT \citep{izacard2022few} mode for faster retrieval. Additionally, we train the retriever with the query side fine-tuning approach which was originally proposed in the ATLAS. This approach is very efficient in model training since it keeps the document encoder frozen while training the parameters corresponding to the query encoder.

\section{Performance Analysis}

In this section, we analyze the performance of the model with respect to the model uncertainty and accuracy scores to address following research questions.
\begin{description}
    \item[\namedlabel{RQ 1}{(RQ 1)}] How does the model confidence differ in scientific tasks when scientific document corpus is used as pretraining and retrieval data?
     \item[\namedlabel{RQ 2}{(RQ 2)}] Does a model have higher confidence in accurate predictions as opposed to inaccurate predictions when provided a scientific document corpus?
\end{description}

To this end, we compute Normalized Predictive Entropy and Semantic Entropy for all the benchmark datasets mentioned in Section \ref{datasets} and compared them across ATLAS (CC + Wiki + S2ORC) and ATLAS (S2ORC) models.
We use 
the logits per sequence from the outermost layer of the decoder in the ATLAS model 
to compute the predictive entropy scores. 
In addition to the token-wise logits, we also use the corresponding text for computing semantic entropy. 
We employ beam search with beam width as $3$ and number of output sequences as $5$. 
\paragraph{RALM finetuned with scientific knowledge as the retrieval data tends to be more confident than the model pretrained only with scientific knowledge}
We report the uncertainty scores of the two ATLAS model variants across three scientific benchmark datasets in Figure~\ref{fig:entropy_accuracy_atlas_sc}.
The results suggest that the uncertainty scores are higher in the ATLAS (S2ORC) model compared to ATLAS (CC+Wiki+S2ORC) for all the benchmark datasets.
We also observe that uncertainty scores differ across different scientific disciplines when measured with semantic entropy as shown in Tables~\ref{fos_entropy_domain} and~\ref{mag_entropy_domain} (see Appendix). For example, ATLAS (S2ORC) model is less confident in all subject areas except Chemistry, Engineering and Psychology related tasks.
This demonstrates that the ATLAS (S2ORC) model is relatively less confident when only scientific knowledge is used as pretraining and retrieval data rather than the mixed of general and scientific training data used in ATLAS (CC+Wiki+S2ORC) model.
On the other hand, general domain data from sources like CC and Wikipedia (news, scientific blogs, articles, and encyclopedia articles) has much more diversity and heterogeneity, which allows for more robust pretraining of the language model~\citep{horawalavithana2022foundation}.

Another reason could be the complexity of the tasks as covered by the scientific benchmark datasets used in the experiments.
FOS and MAG datasets cover more than 10 scientific disciplines and test the ability of the models to understand diverse scientific domains and disciplines\footnote{FoS tasks include instructions from following domains;  Materials science,  Economics,  Chemistry,  Medicine,  Psychology,  Geography,  Geology,  Political science,  Engineering,  Philosophy,  Sociology,  Physics,  Computer science,  Law,  History,  Biology,  Agricultural and Food sciences,  Environmental science,  Business,  Education,  Art,  Linguistics,  Mathematics}.
MMLU assess how well models can understand and use knowledge to answer questions from diverse scientific disciplines in 
Biology,
Chemistry,
Computer science,
Mathematics,
Physics,
etc.
Retrieving scientific knowledge from two or more different disciplines might help the Reader component to identify the connections between different ideas and concepts, and solve problems more effectively.
Having a Reader trained with a mix of general and scientific datasets would allow it to better understand and generate text on a variety of topics. The general datasets would provide the Reader with a broad understanding of the world, while the scientific datasets would give it the knowledge to understand and generate text on more specific topics. This would make the Reader more versatile and useful for a variety of tasks.


\begin{figure*}[htbp]
    \centering
    \begin{subfigure}[t]{0.5\textwidth}
        \centering
        \includegraphics[width=\textwidth]{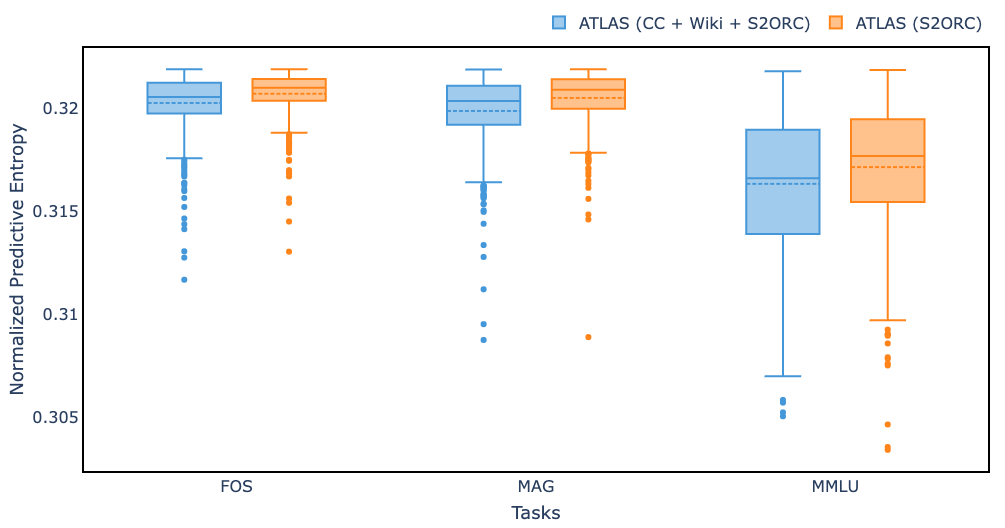}
        \caption{Normalized Predictive Entropy}
        \label{fig:norm_entropy_tasks}
    \end{subfigure}%
    ~ 
    \begin{subfigure}[t]{0.5\textwidth}
        \centering
        \includegraphics[width=\textwidth]{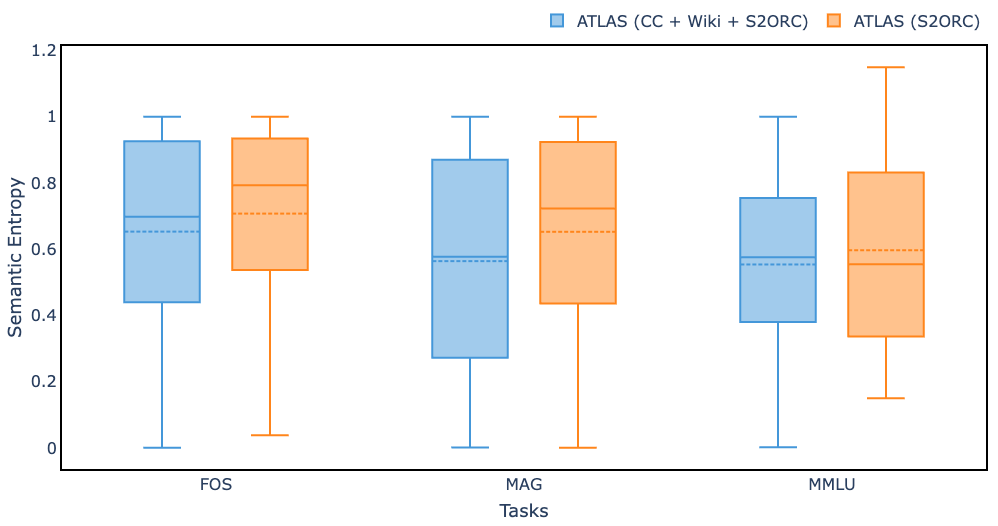}
        \caption{Semantic Entropy}
        \label{fig:sem_entropy_tasks}
    \end{subfigure}
    \caption{Uncertainty scores across FOS, MAG and MMLU scientific benchmarks}
    \label{fig:entropy_accuracy_atlas_sc}
\end{figure*}

\begin{figure*}[htbp]
    \centering
    \begin{subfigure}[t]{0.5\textwidth}
        \centering
        \includegraphics[width=\textwidth]{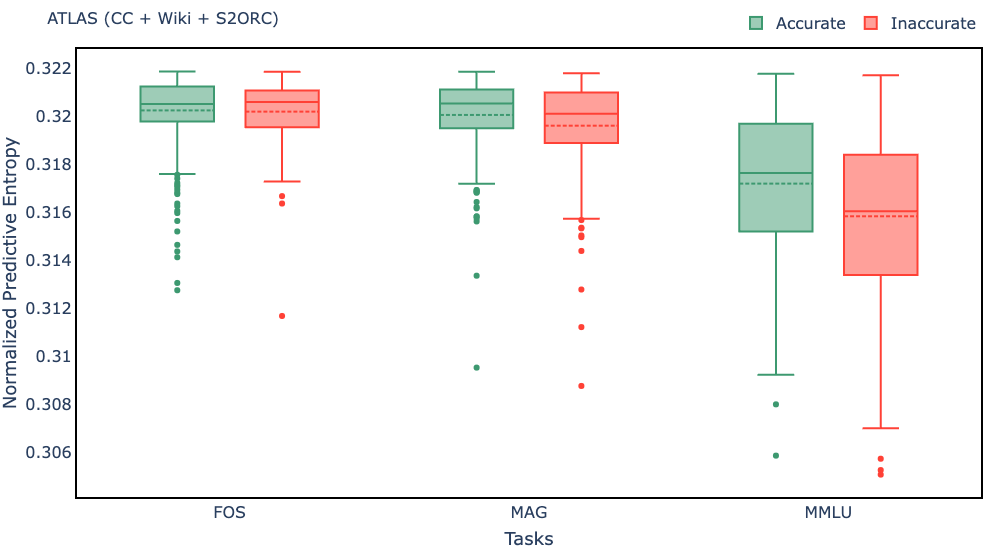}
        \caption{ATLAS (CC+ Wiki + S2ORC): Predictive Entropy}
        \label{fig:ent_accuracy_atlas}
    \end{subfigure}%
    ~ 
    \begin{subfigure}[t]{0.5\textwidth}
        \centering
        \includegraphics[width=\textwidth]{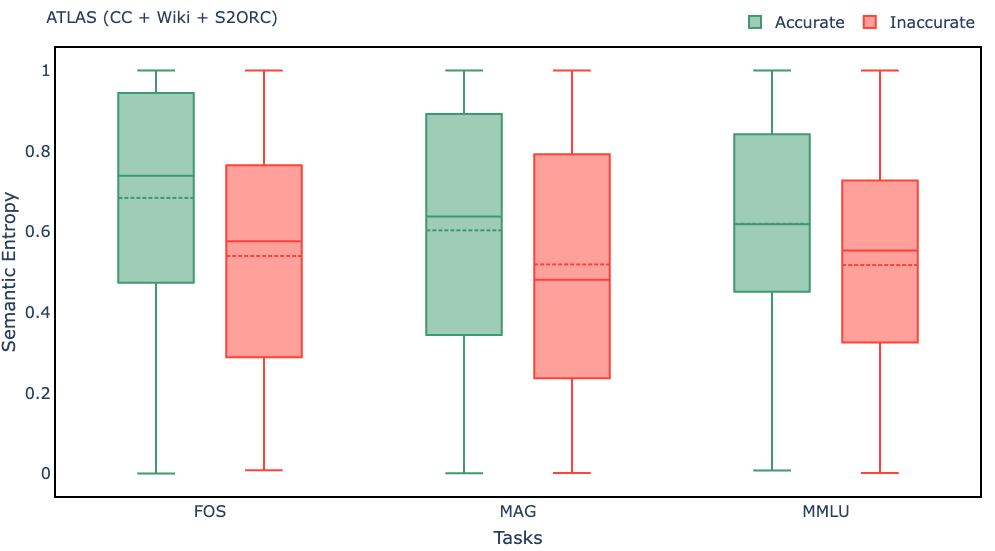}
        \caption{ATLAS (CC+ Wiki + S2ORC): Semantic Entropy}
        \label{fig:se_accuracy_atlas}
    \end{subfigure}



    \begin{subfigure}[t]{0.5\textwidth}
        \centering
        \includegraphics[width=\textwidth]{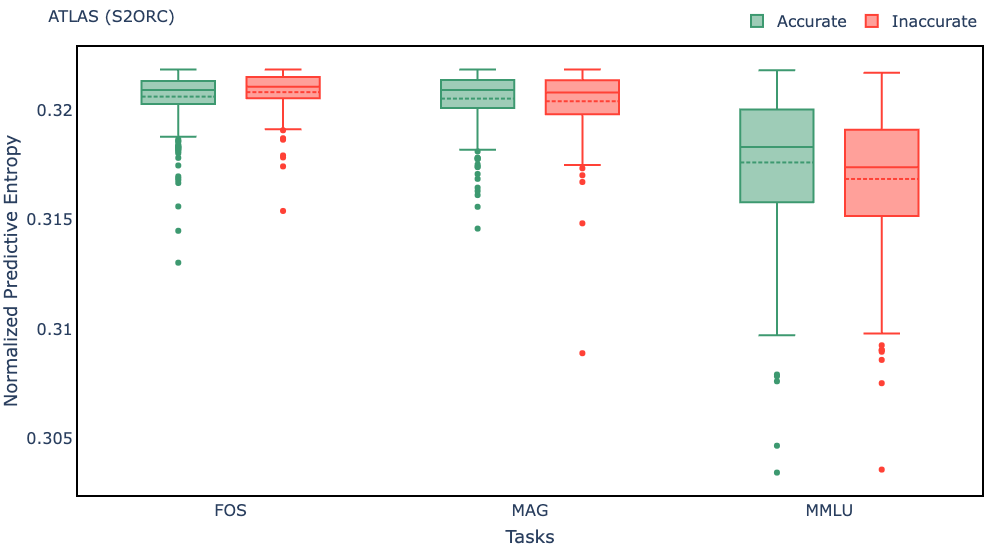}
        \caption{ATLAS (S2ORC): Predictive Entropy}
        \label{fig:ent_accuracy_atlas_science}
    \end{subfigure}%
    ~ 
    \begin{subfigure}[t]{0.5\textwidth}
        \centering
        \includegraphics[width=\textwidth]{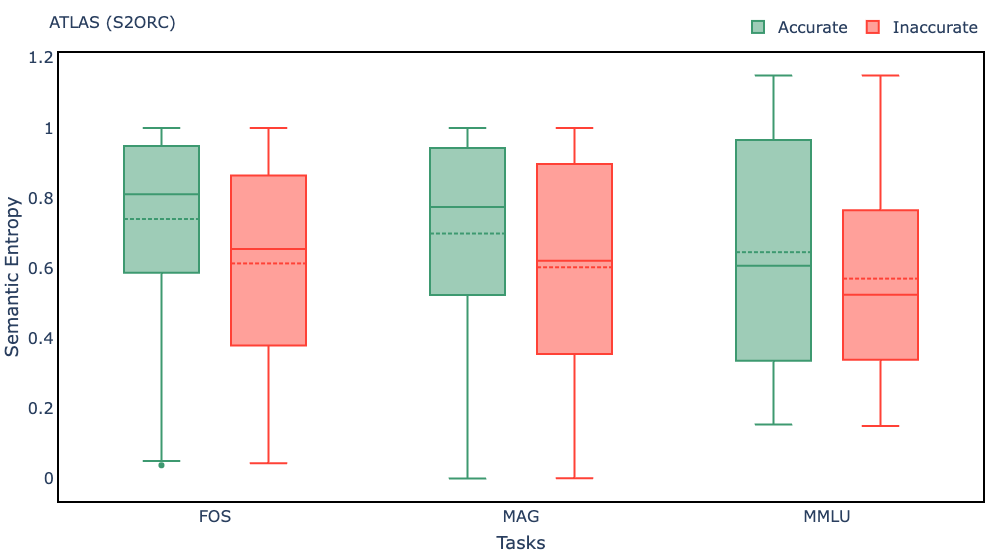}
        \caption{ATLAS (S2ORC): Semantic Entropy}
        \label{fig:se_accuracy_atlas_science}
    \end{subfigure}
    \caption{Uncertainty scores as grouped by the accurate and inaccurate predictions}
    \label{fig:entropy_accuracy_atlas_sc}
\end{figure*}

\paragraph{RALMs make inaccurate predictions more confidently than making accurate predictions.}
In this section, we compare the uncertainty scores when the models make accurate and inaccurate predictions (as shown in Figures~\ref{fig:ent_accuracy_atlas}~-~\ref{fig:se_accuracy_atlas_science}).
The results suggest that the uncertainty scores are higher for accurate predictions in comparison to the inaccurate predictions.
To state it differently, RALMs probability estimates are not in good agreement with the actual probability of the answer being correct. 

One can argue this might be due to the distributional characteristics of the accurate and inaccurate predictions made by the models.
However, the majority of predictions ($>80\%$) in FOS and MAG tasks are accurate, while the majority of MMLU predictions ($>60\%$) are inaccurate.
Despite these distributional differences, both models are less confident while predicting the accurate answers and vice versa.
We believe this is mainly due to LLMs/RALMs not calibrating well to the downstream tasks.

Our findings are consistent with the previously reported calibration issues~\citep{jiang2021can} in standard large language models (LLMs) such as T5, BART, and GPT-2.
We demonstrate that these issues continue to exist in RALMs when applied to scientific tasks, and that scientific knowledge provided either as pretraining or retrieval data does not help the model to alleviate the calibration issues.
Finding calibration methods for RALMs, especially when applied to scientific tasks, remains a work in progress.
In the future, we also plan to conduct experiments on larger scale models to analyze the impact of scale on UQ measures.


\section{Conclusions}
In this paper, we conduct a comprehensive evaluation of uncertainties in retrieval augmented language models for scientific tasks.
We leverage ATLAS as our test models with SciRepEval and MMLU as benchmarks. We compare the general domain ATLAS model with that of pretrained from scratch on scientific literature (S2ORC). 
Experiments indicate that the model finetuned with scientific knowledge as the retrieval data tends to be more confident than the model pretrained only with scientific knowledge.
We also observed calibration issues in RALMs that are more confident in inaccurate predictions compared to accurate ones in science tasks.

\section*{Acknowledgements}
This work was supported by the NNSA Office of Defense Nuclear Nonproliferation Research and Development, U.S. Department of Energy, and Pacific Northwest National Laboratory, which is operated by Battelle Memorial Institute for the U.S. Department of Energy under Contract DE-AC05–76RLO1830. This article has been cleared by PNNL for public release as PNNL-SA-191164. 

\bibliography{main}
\bibliographystyle{main}

\appendix
\section*{ Appendix A: Domain wise Performance Analysis} \label{appendix}


Table \ref{fos_entropy_domain} summarizes domain wise model uncertainty scores (Normalized Predictive Entropy and Semantic Entropy) for FOS benchmark dataset. 
\begin{table}[!h]
\caption{Summary of domain wise uncertainty scores for FOS dataset}
\label{fos_entropy_domain}
\begin{center}
\scalebox{0.85}
{
\begin{tabular}{|l|ll|ll|}
\hline
            & \multicolumn{2}{c|}{Normalized Predictive Entropy}         & \multicolumn{2}{c|}{Semantic Entropy}           \\ \hline
            & \multicolumn{1}{l|}{ATLAS (CC + Wiki + S2ORC)} & ATLAS (S2ORC) & \multicolumn{1}{l|}{ATLAS (CC + Wiki + S2ORC)} &  ATLAS (S2ORC) \\ \hline
Art         & \multicolumn{1}{l|}{0.3192}     & 0.3197        & \multicolumn{1}{l|}{0.5945}     & 0.7015        \\ \hline
Biology     & \multicolumn{1}{l|}{0.3204}     & 0.3208        & \multicolumn{1}{l|}{0.6244}     & 0.7156        \\ \hline
Business    & \multicolumn{1}{l|}{0.3195}     & 0.3205        & \multicolumn{1}{l|}{0.6258}     & 0.6680        \\ \hline
Chemistry   & \multicolumn{1}{l|}{0.3206}     & 0.3203        & \multicolumn{1}{l|}{0.7018}     & 0.6847        \\ \hline
Economics   & \multicolumn{1}{l|}{0.3192}     & 0.3209        & \multicolumn{1}{l|}{0.5557}     & 0.6366        \\ \hline
Engineering & \multicolumn{1}{l|}{0.3209}     & 0.3208        & \multicolumn{1}{l|}{0.7296}     & 0.6551        \\ \hline
Geology     & \multicolumn{1}{l|}{0.3211}     & 0.3210        & \multicolumn{1}{l|}{0.7998}     & 0.8282        \\ \hline
History     & \multicolumn{1}{l|}{0.3195}     & 0.3201        & \multicolumn{1}{l|}{0.7301}     & 0.8275        \\ \hline
Mathematics & \multicolumn{1}{l|}{0.3200}     & 0.3208        & \multicolumn{1}{l|}{0.5410}     & 0.7076        \\ \hline
Medicine    & \multicolumn{1}{l|}{0.3203}     & 0.3209        & \multicolumn{1}{l|}{0.5793}     & 0.6784        \\ \hline
Philosophy  & \multicolumn{1}{l|}{0.3198}     & 0.3207        & \multicolumn{1}{l|}{0.6012}     & 0.6574        \\ \hline
Physics     & \multicolumn{1}{l|}{0.3209}     & 0.3208        & \multicolumn{1}{l|}{0.7314}     & 0.7824        \\ \hline
Psychology  & \multicolumn{1}{l|}{0.3212}     & 0.3209        & \multicolumn{1}{l|}{0.7990}     & 0.7272        \\ \hline
Sociology   & \multicolumn{1}{l|}{0.3201}     & 0.3207        & \multicolumn{1}{l|}{0.5340}     & 0.6382        \\ \hline
\end{tabular}
}
\end{center}

\end{table}

Table \ref{mag_entropy_domain} summarizes domain wise model uncertainty scores (Normalized Predictive Entropy and Semantic Entropy) for MAG benchmark dataset. 
\begin{table}[H]
\caption{Summary of domain wise uncertainty scores for MAG dataset}
\label{mag_entropy_domain}
\scalebox{0.85}{
\begin{tabular}{|l|ll|ll|}
\hline
            & \multicolumn{2}{c|}{\text{Normalized Predictive Entropy}}    & \multicolumn{2}{c|}{Semantic Entropy}                          \\ \hline
            & \multicolumn{1}{l|}{ATLAS (CC + Wiki + S2ORC)} & ATLAS (S2ORC) & \multicolumn{1}{l|}{ATLAS (CC + Wiki + S2ORC)} & ATLAS (S2ORC) \\ \hline
Art         & \multicolumn{1}{l|}{0.3192}                    & 0.3203        & \multicolumn{1}{l|}{0.5014}                    & 0.7153        \\ \hline
Biology     & \multicolumn{1}{l|}{0.3198}                    & 0.3201        & \multicolumn{1}{l|}{0.4212}                    & 0.4862        \\ \hline
Business    & \multicolumn{1}{l|}{0.3203}                    & 0.3203        & \multicolumn{1}{l|}{0.5155}                    & 0.5405        \\ \hline
Chemistry   & \multicolumn{1}{l|}{0.3201}                    & 0.3210        & \multicolumn{1}{l|}{0.5329}                    & 0.6656        \\ \hline
Economics   & \multicolumn{1}{l|}{0.3197}                    & 0.3208        & \multicolumn{1}{l|}{0.6946}                    & 0.7213        \\ \hline
Engineering & \multicolumn{1}{l|}{0.3203}                    & 0.3207        & \multicolumn{1}{l|}{0.3889}                    & 0.5852        \\ \hline
Geology     & \multicolumn{1}{l|}{0.3204}                    & 0.3204        & \multicolumn{1}{l|}{0.6156}                    & 0.7894        \\ \hline
History     & \multicolumn{1}{l|}{0.3190}                    & 0.3205        & \multicolumn{1}{l|}{0.5258}                    & 0.6821        \\ \hline
Mathematics & \multicolumn{1}{l|}{0.3195}                    & 0.3209        & \multicolumn{1}{l|}{0.5245}                    & 0.6878        \\ \hline
Medicine    & \multicolumn{1}{l|}{0.3203}                    & 0.3206        & \multicolumn{1}{l|}{0.6384}                    & 0.7624        \\ \hline
Philosophy  & \multicolumn{1}{l|}{0.3180}                    & 0.3200        & \multicolumn{1}{l|}{0.4936}                    & 0.4783        \\ \hline
Physics     & \multicolumn{1}{l|}{0.3211}                    & 0.3211        & \multicolumn{1}{l|}{0.8701}                    & 0.8765        \\ \hline
Psychology  & \multicolumn{1}{l|}{0.3208}                    & 0.3204        & \multicolumn{1}{l|}{0.6229}                    & 0.5840        \\ \hline
Sociology   & \multicolumn{1}{l|}{0.3195}                    & 0.3200        & \multicolumn{1}{l|}{0.5546}                    & 0.5658        \\ \hline
\end{tabular}
}
\end{table}
\section*{Appendix B: Human-AI Reasoning with Uncertainty Quantification} \label{appendix_widgets}

\begin{figure*}[h!]
    \centering
    \begin{subfigure}[t]{0.95\textwidth}
        \centering
        \includegraphics[scale=0.45]{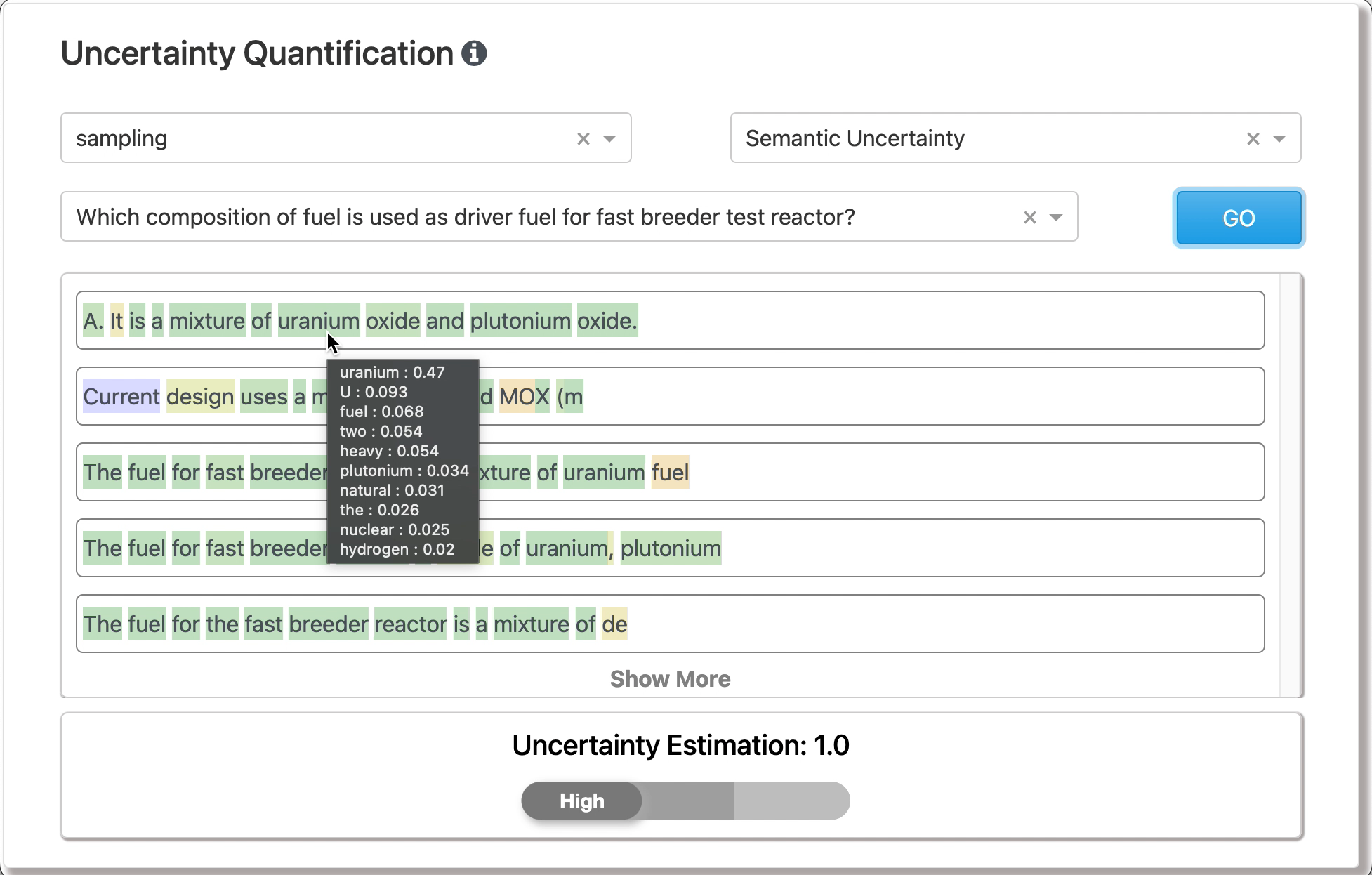}
        \caption{Visualizing token-level probabilities}
        \label{fig:uq_widget_a}
    \end{subfigure}%
    \\
    \begin{subfigure}[t]{0.95\textwidth}
        \centering
       \includegraphics[scale=0.45]{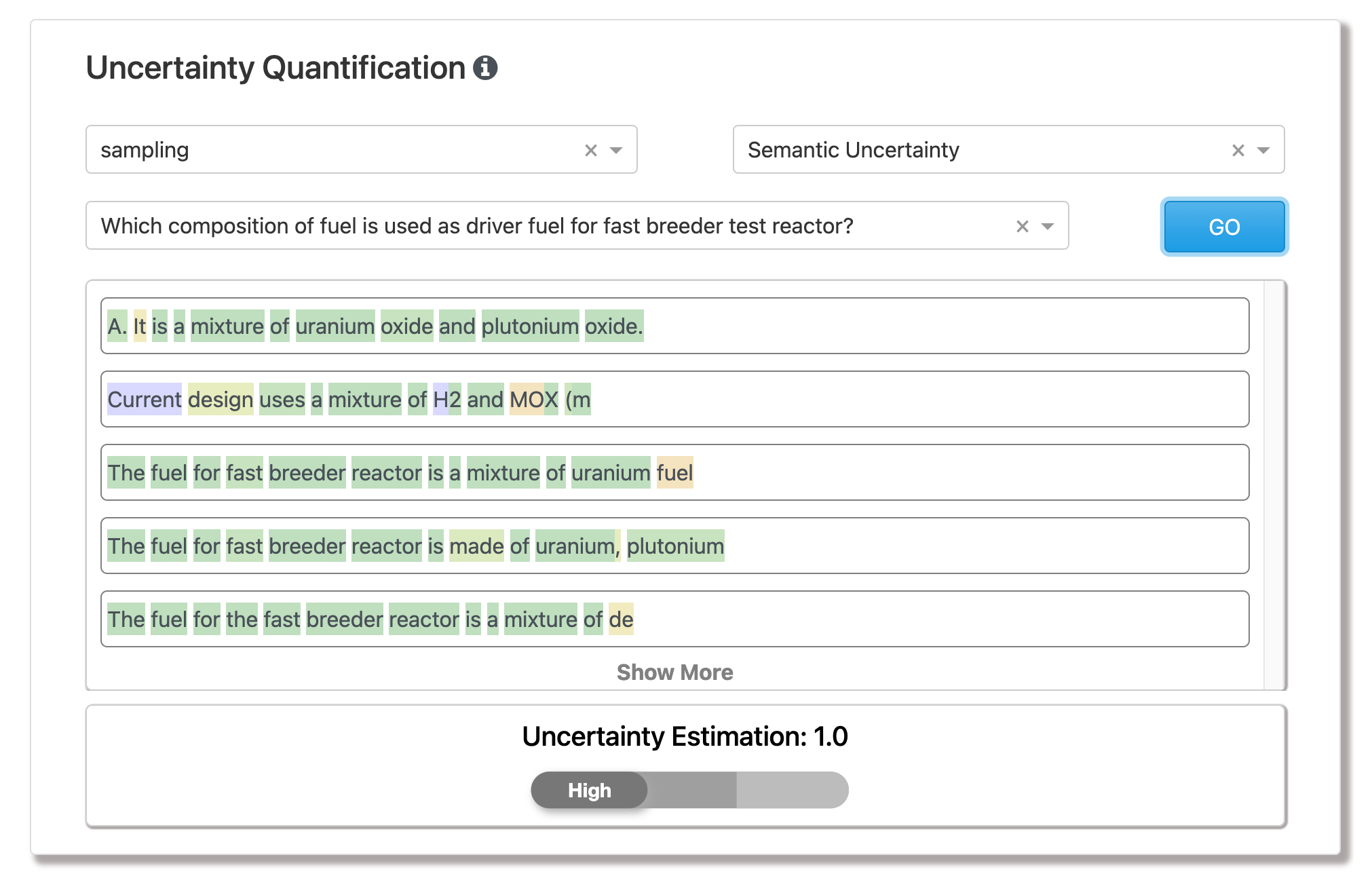}
        \caption{Visualizing multiple generated sequences}
        \label{fig:uq_widget_b}
    \end{subfigure}
    \caption{Human-AI Reasoning Dashboard with Uncertainty Quantification}
    \label{fig:uq_widget}
\end{figure*}

Figure~\ref{fig:uq_widget} shows the uncertainty quantification widgets that are built using Plotly Dash, a python-based framework used for rapid prototyping and analytic tool development. 
The dashboard supports three different types of text generation methods:
\begin{enumerate}
    \item Sampling with temperature: This method randomly picks the next token from a set of high-probability tokens
    \item Nucleus (Top-p) Sampling: This method chooses from the smallest possible set of tokens whose cumulative probability exceeds the probability p
    \item Beam Search: This method keeps the most likely number of beams for hypotheses at each time step and eventually chooses the hypothesis that has the overall highest probability.
\end{enumerate}

\end{document}